\documentclass[]{ceurart}
\usepackage{subcaption}
\usepackage{color,soul}
\usepackage[most]{tcolorbox}

\definecolor{com0}{RGB}{102, 197, 204}
\definecolor{com1}{RGB}{246, 207, 113}
\definecolor{com2}{RGB}{248, 156, 116}
\definecolor{com3}{RGB}{220, 176, 242}
\definecolor{com4}{RGB}{135, 197, 95}
\definecolor{com5}{RGB}{158, 185, 243}
\definecolor{partyafd}{RGB}{0, 158, 224}
\definecolor{partygreens}{RGB}{0, 84, 55}
\definecolor{partycdu}{RGB}{0, 0, 0}
\definecolor{partycsu}{RGB}{197, 228, 237}
\definecolor{partyspd}{RGB}{227, 0, 15}
\definecolor{partyfdp}{RGB}{255, 226, 9}

\begin{document}
\copyrightyear{2025}
\copyrightclause{Copyright for this paper by its authors.
  Use permitted under Creative Commons License Attribution 4.0
  International (CC BY 4.0).}

\conference{In: R. Campos, A. Jorge, A. Jatowt, S. Bhatia, M. Litvak (eds.): Proceedings of the Text2Story'25 Workshop, Lucca (Italy), 10-April-2025}

\title{Automated Identification of Competing Narratives in Political Discourse on Social Media}

\author[1]{Sergej Wildemann}[%
orcid={0000-0001-9216-4253},
email=wildemann@l3s.de,
]
\address[1]{L3S Research Center, Leibniz Universität Hannover, Hannover, Germany}
\cormark[1]
\author[1]{Erick Elejalde}[%
orcid={0000-0002-4755-1606},
email=elejalde@l3s.de,
]
\cortext[1]{Corresponding author.}

\begin{abstract}
  Social media platforms have become central to shaping political discourse, serving as arenas where narratives form and evolve, influencing public opinion. Identifying and analyzing these narratives, particularly when they compete across different political ideologies, is crucial for understanding the dynamics of modern political communication. This paper presents an unsupervised framework for identifying and characterizing competing narratives in political discourse on social media, focusing on German politicians’ tweets. The framework employs a multi-stage pipeline that integrates natural language processing techniques such as topic modeling, event detection, and event linking. By forming data into coherent stories and uncovering the distinct perspectives of user communities, the system is able to detect the key competing narratives, highlighting the divergent framings and conflicts surrounding trending political topics. Two case studies on polarizing political issues demonstrate the efficacy of the methodology, showcasing its ability to uncover and analyze divergent viewpoints. The findings contribute to the broader understanding of how narratives propagate within the digital public sphere and offer insights for policymakers, social media platforms, and researchers interested in monitoring political discourse.
\end{abstract}


\maketitle
\section{Introduction}

Social media has become a central arena for political discourse, shaping public opinion. Platforms like X (formerly Twitter) are key outlets for politicians to communicate directly with citizens \cite{gilardi2022social}.
Social media allows politicians to craft and directly disseminate narratives that resonate with their followers and political base \cite{silva2022politicians}.
These narratives—structured interpretations of events and issues—often compete with those from other groups, reflecting differing political ideologies and perspectives on pressing societal matters \cite{jing2021characterizing}.
This can be seen, for example, in contrasting framings of events related to health crises \cite{jing2021characterizing} or climate change \cite{ross2019internet}, where some groups emphasize a scientific perspective while others focus on belief-based interpretations \cite{reiter2024framing}. 
Understanding how these competing narratives emerge, evolve, and propagate is essential for capturing the dynamics of contemporary political communication and addressing societal challenges such as polarization, misinformation, and the shaping of public opinion. 

Our work contributes to the development of computational narrative framing analysis \cite{reiter2024computational}. 
Specifically, this paper addresses the challenge of automatically identifying and analyzing competing narratives within the political discourse of German politicians on Twitter. 
For this, we propose a novel framework grounded in natural language processing (NLP) and clustering techniques. 
By focusing on political narratives in the social media context, we tap into a naturally polarized environment with frames that develop across multiple documents (e.g., tweets) and in a (potentially) collaborative way involving multiple sources.
This analysis can provide insights into how political actors shape online debate and how competing viewpoints may influence societal understanding of political events.

This research area holds promise for fostering evidence-based approaches to societal challenges. For example, promoting media literacy and critical thinking through exposure to diverse viewpoints can empower individuals to evaluate information more critically, identify biases, and make informed decisions \cite{axelrod2021preventing}. Moreover, understanding competing narratives can facilitate constructive dialogue by highlighting shared values and differences, enabling collaborative problem-solving. The automatic extraction and analysis of competing narratives can advance efforts to combat misinformation, improve public understanding, and support more critical engagement with media content.

\section{Background and Related Work}

Narrative studies have a long history across various disciplines, including literary studies, psychology, and sociology.
Narratives are typically defined as representations of a connected succession of events \cite{mishler1995models} and often blend reality and fiction to influence choices \cite{page2013stories, jones2010narrative}.
They serve as systems of interrelated stories that may include competing storylines \cite{page2013stories}.
Various models of narrative analysis have been proposed, each focusing on different aspects \cite{mishler1995models}.
For example, models can focus on temporal ordering, textual coherence, or the social functions of narratives.
Models concerned with textual coherence examine how narratives are structured and organized to create a cohesive and understandable story.
Conversely, models based on references and temporal order analyze actual events and their sequencing rather than just their textual representation \cite{mishler1995models}.
Finally, models focusing on social functions investigate how narratives are used in social contexts to persuade, entertain, or inform a target audience \cite{mishler1995models}.
This theoretical work lays the foundations for understanding narratives and developing practical computational methods for their analysis.
Our approach combines various models to automatically identify narratives that are temporally and semantically coherent while considering their social functions, especially among political actors on social media.

A crucial aspect of computational narrative understanding is the automatic extraction of narrative elements from text data.
This includes identifying actors, actions, events, settings, and the relationships between them \cite{keith2023survey, mani2022computational, metilli2019steps}. 
For example, researchers have developed methods for identifying events, linking them based on temporal and causal relationships, and representing event-based narrative structures in different domains \cite{keith2023survey}. 
One promising representation approach employs a route map metaphor to visualize the events and storylines within a narrative, thereby facilitating understanding of the narrative's central themes \cite{keith2021narrative}. 
These algorithms also assist in extracting events and storylines based on coherence and coverage \cite{keith2021narrative,DBLP:conf/acl/YuJDSY20}.
Furthermore, techniques for semantic role labeling and semantic graphs can effectively improve the analysis of narrative frames by extracting detailed information about the roles of different actors and actions within a story \cite{reiter2024framing, metilli2019steps, otmakhova-etal-2024-media}.
These building blocks are essential because reliable computational techniques enable analysts to scale their investigation to large volumes of narrative data, allowing for a deeper understanding of narrative structures and their impact.

Framing understanding is another crucial aspect in shaping the reader's perception of a narrative.
Computational framing analysis aims to automatically detect framing devices and understand how they are used to influence interpretation \cite{reiter2024computational}.
Previous works have recognized specific linguistic patterns and rhetorical techniques that signal framings, such as word choice, metaphors, and the presentation of evidence \cite{liu2019detecting, otmakhova-etal-2024-media}.
Frames can also be studied through sentiment analysis, where determining the emotional tone expressed toward different actors and events reveals evoked emotions and attitudes in the reader \cite{gitari2015lexicon, pitropakis2020monitoring}.
Analyzing how specific frames are associated with various actors and concepts within a narrative helps uncover potential biases and perspectives \cite{hamborg2019automated, wildemann2023migration}.
Computational framing analysis has shown to be effective in distinguishing between framing across different sources, such as conspiracy versus mainstream media, and in unveiling media bias  \cite{hamborg2019automated, reiter2024computational}.
Nonetheless, challenges remain in consistently capturing framing information across narrative chains %
\cite{mendelsohn2021modeling}.
We adopt a computational narrative framing approach that examines the framing based on the structure of the narrative (e.g., the inclusion/exclusion of events by specific sources). 
However, more content-driven methods of framing analysis (e.g., sentiment analysis) can complement our methodology to provide additional insights. 

Finally, studying competing narratives is essential for understanding how different perspectives and interpretations of events emerge and spread.
This could help to monitor community interests, to counter propaganda campaigns, and to model the spread of (mis-)information \cite{hamborg2019automated1, reiter2024computational}.
The task of computational approaches in this area focuses on identifying and tracking competing narratives' evolution over time and understanding the factors that drive their dynamics \cite{bandeli2020framework, schmid2017predicting}.
For this, proposed methods should distinguish between different narratives based on variations in actors, actions, framing devices, and the overall message conveyed.
Competing narratives can then be tracked by analyzing changes in language use, sentiment, framing devices, and the prominence of different actors and actions over time.
Investigating how external events, cultural shifts, and other narratives influence the emergence and spread of competing narratives allows us to understand their driving factors.

\section{Dataset}
For our study, we focus on the narratives within the political discourse of German politicians on Twitter/X.
Before collecting the corresponding tweets, the relevant politicians and their Twitter accounts must first be identified.
For this task, we leverage the structured database Wikidata, which provides information on politicians' affiliations with political parties and their country of citizenship.

We filter Wikidata entities by the {\em occupation} property matching \enquote{politician}.
Subsequently, the {\em country of citizenship} is utilized to filter German politicians.
Their party affiliation is provided by the {\em member of political party} property.
However, this property may list multiple parties, as politicians may change their affiliation over time.
In certain instances, the current party is marked by a qualifier.
Otherwise, we consider the first party listed, ignoring any affiliation with an end date.
Extracting Twitter usernames from Wikidata with the above filters yielded 1,324 accounts belonging to 1,300 politicians.

Using Twitter's API, we collected all tweets of these politicians from January 1st, 2022, to June 24th, 2023.
The resultant dataset comprised 189,850 tweets from 786 accounts, excluding retweets, replies, and quotes.
For presentation purposes, we translated the tweets into English using the NLLB model~\cite{team2022NoLL}.
Otherwise, all the analyses are conducted on the original content (i.e., original language).

\section{Methodology}
\begin{figure}[ht]
    \centering
    \includegraphics[width=0.98\linewidth,trim={13mm 13mm 13mm 13mm},clip]{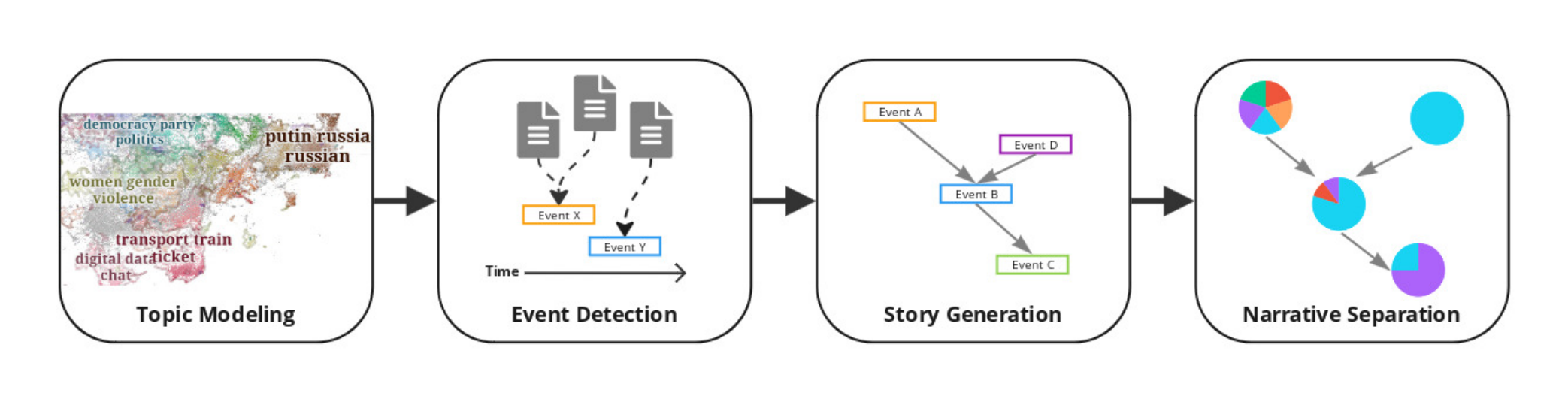}
    \caption{Competing Narrative Extraction Pipeline: Separation of user viewpoints within stories of a topic.}
    \label{fig:pipeline}
\end{figure}

We operationalize narratives as semantically and temporarily coherent stories represented by structured interpretations of events and issues, providing meaning and context to political communication.
Starting with the social media posts, we sought to identify these stories and the competing narratives of different user groups within them.
Given a single tweet's brevity and limited scope, we follow a corpus-level approach to identify narratives rather than trying to extract them from individual tweets.

Our methodology consist of a multi-stage pipeline (see Fig.~\ref{fig:pipeline}).
After identifying general topics within the data, we then detect events that form the building blocks of our stories.
Events are linked into coherent stories from which we extract different viewpoints, which we refer to as competing narratives.

Note that, during the narrative extraction process, we disregard tweet authors' political affiliations, as this information is unique to our dataset and may not be available in other contexts.
However, it will be used during the evaluation to assess the effectiveness of our method to identify competing narratives.

We are publicly releasing the full source code for our data processing pipeline\footnote{\url{https://github.com/fjen/competing-narratives}} to promote open science and ensure reproducibility.

\subsection{Topic Modeling}
Given the diversity of topics in the dataset, the first step is to identify individual topics and remove noise.
An unsupervised topic model is trained to group documents based on their semantic similarity.
Here, we choose an embedding-based approach, as it is more robust to the short and noisy nature of tweets than traditional topic models like LDA.
In particular, we rely on the modular BERTopic framework~\cite{grootendorst2022bertopic}, which essentially clusters documents based on their semantic embeddings and generates text representations for each topic.
Notably, we use Sentence-BERT with the \texttt{paraphrase-multilingual-MiniLM-L12-v2} model \cite{reimers-2019-sentence-bert} for embedding generation and set a minimum cluster size of 500 documents to avoid overly fine-grained topics.

\begin{table}
\caption{Excerpt of Topics Identified in the Dataset}
\label{tab:topics}
\tiny
\renewcommand{\arraystretch}{0.8}
\setlength{\tabcolsep}{2pt} 

\begin{tabular}[t]{cccl}
\toprule
\textbf{Topic} & \textbf{Documents} & \textbf{Users} & \textbf{Keywords} \\
\midrule
0 & 11516 & 558 & putin, russia, russian, war, ukraine \\
1 & 5898 & 519 & going, good, morning, day, know \\
2 & 5237 & 505 & transport, train, ticket, traffic, mobility \\
3 & 4477 & 485 & ukraine, tanks, war, weapons, ukrainian \\
4 & 3990 & 439 & vaccination, vaccine, corona, pandemic, coronavirus \\
5 & 3859 & 446 & tax, inflation, debt, budget, discharge \\
6 & 3789 & 460 & children, education, child, wage, minimum \\
7 & 3697 & 448 & women, gender, violence, men, queer \\
8 & 3360 & 462 & berlin, great, concert, festival, today \\
9 & 3165 & 420 & anti, israel, semitism, anti semitism, victims \\
10 & 3084 & 345 & digital, data, chat, chat control, control \\
11 & 3026 & 416 & energy, gas, renewable, electricity, price \\
\bottomrule
\end{tabular}\hspace{3mm}
\begin{tabular}[t]{cccl}
\toprule
\textbf{Topic} & \textbf{Documents} & \textbf{Users} & \textbf{Keywords} \\
\midrule
12 & 2836 & 413 & climate, climate protection, protection, climate change, change \\
13 & 2819 & 431 & european, eu, europe, european parliament, parliament \\
14 & 2506 & 309 & water, food, forest, agriculture, nature \\
15 & 2303 & 435 & greens, green, spd, fdp, party \\
16 & 2142 & 427 & qatar, football, congratulations, cup, world cup \\
17 & 1974 & 279 & iran, regime, iranian, prison, prisoners \\
18 & 1932 & 307 & nuclear, nuclear power, power, power plants, plants \\
19 & 1788 & 388 & democracy, party, politics, left, reform \\
20 & 1734 & 381 & chancellor, scholz, merkel, olaf, olaf scholz \\
21 & 1606 & 258 & 39 39, sk, 39, people, war \\
22 & 1554 & 258 & china, chinese, taiwan, xi, beijing \\
23 & 1472 & 297 & asylum, refugees, migration, immigration, refugee \\
\bottomrule
\end{tabular}
\end{table}

After disregarding outliers (i.e., not clustered tweets), we identify 57 topics in the dataset containing 108,063 documents.
A significant portion of the documents are labeled as outliers (43\%), which is expected given the nature of the data.
As shown in Table~\ref{tab:topics}, the remaining documents are grouped into topics that successfully capture the main discussion strands during the observation period with examples like the war in Ukraine, the COVID-19 pandemic, or Europe's energy crisis.

\subsection{Event Detection}
We define an event as a set of documents that discuss the same issue in close temporal proximity \cite{DBLP:conf/acl/YuJDSY20}.
This definition implies that documents within an event share similar identifying information, such as keywords, actors, and/or locations.
However, semantic similarity alone might lead to similar but distinct discussions being grouped together.
Thus, the temporal aspect is crucial for clearly separating events.

For the extraction of events, we borrow parts of the proposed Multi-TimeLine Summarization (MTLS) approach by \citeauthor{DBLP:conf/acl/YuJDSY20}~\cite{DBLP:conf/acl/YuJDSY20}.
Although we use the same definition of events, their work is centered on (larger) news articles, whereas we focus on (shorter) tweets.
Consequently, rather than relying on individual sentences, we utilize the entire tweet as the fundamental unit of analysis.

Tweets are clustered into events using the affinity propagation (AP) algorithm~\cite{frey2007clustering}.
It operates on an affinity matrix $S_1$, where $S_1(i, j)$ represents the similarity between tweets $i$ and $j$ based on a linear combination of their semantic similarity and temporal proximity:

\begin{equation}
  \label{eq:event-affinity}
  S_1(i, j) = \alpha \cdot S_{temp}(i, j) + (1 - \alpha) \cdot S_{cos}(i, j)
\end{equation}

\noindent where $S_{cos}$ is the cosine similarity of the tweet embeddings and $S_{temp}$ is the temporal similarity defined by an exponential function (Eq.~\ref{eq:event-temporal}) with a decay factor $\lambda$:

\begin{equation}
\label{eq:event-temporal}
S_{temp}(i, j) = \frac{1}{\exp^{\lambda \cdot |t_i - t_j|}}
\end{equation}

Our time unit is in days and we set $\lambda$ to 0.05 as suggested by \citeauthor{DBLP:conf/acl/YuJDSY20}.
Also, $\alpha$ is set to 0.4 to balance the similarity measures.
For example, we obtain 155 events for the \emph{"energy"} topic (Topic 11 in Table \ref{tab:topics}).

\subsection{Story Generation}
Next, we group events into coherent stories, creating sequences of events that are meaningfully related.
Since we are working at the document level and not at the sentence level, we cannot use a procedure similar to MTLS for event linking, which is based on sentence co-occurrence.
Our approach is based on an event similarity graph where we filter the most significant edges to partition the events into stories.

\paragraph{Event Similarity Measure}
Since each event contains multiple tweets, we must define the semantic %
similarity measure on the inter-event level.
The AP algorithm used to cluster those tweets also returns documents at the cluster centers as representatives of the events~\cite{DBLP:conf/acl/YuJDSY20}.
However, these centers may not cover the full breadth and depth of the event discussion.
The same applies if calculating centroids of the tweet embeddings.
Thus, we calculate semantic event similarity using the Sliced Wasserstein Distance (SWD) of the tweet embeddings within event pairs.
Unlike cosine similarity, which focuses on the angular difference between pairs of document embeddings, SWD measures the distance between entire embedding distributions.
It does this by slicing the multidimensional embedding space along various directions, computing Wasserstein distances in these 1D projections, and then averaging the results.
Additionally, given that we average $15.9$ tweets per event, SWD is computationally more efficient than calculating pairwise document distances.
Finally, the SWD of the two distributions is transformed to the semantic similarity via the exponential decay function: $S_{sem} = \exp^{-\text{SWD}(e_1, e_2)}$.

Similarly to event detection, an affinity matrix $S_2$ is created based on the similarity of event embeddings and a penalty term, temporal distance, to avoid connecting temporally distant events, which in turn could lead to improbable stories spanning the entire time range.
We define the event similarity $S_2(e_1, e_2)$ between a pair of events $e_1$ and $e_2$ as:

\begin{equation}
  \label{eq:story-affinity}
  S_2(e_1, e_2) = S_{sem}(e_1, e_2) \cdot {\exp^{-\gamma \cdot |t_{e_1} - t_{e_2}|}}
\end{equation}

\noindent where $S_{sem}$ is the similarity of the event embeddings, and $t_{e_i}$ is the timestamp of the representative document within an event.
The temporal penalty decreases the similarity score as the time difference between the events increases, with the decrease rate controlled by the $\gamma$ parameter.
We set $\gamma = 0.01$.

\paragraph{Forming Storylines}
\begin{figure}[ht]
    \centering
    \includegraphics[width=\linewidth,trim={10mm 16mm 8mm 27mm },clip]{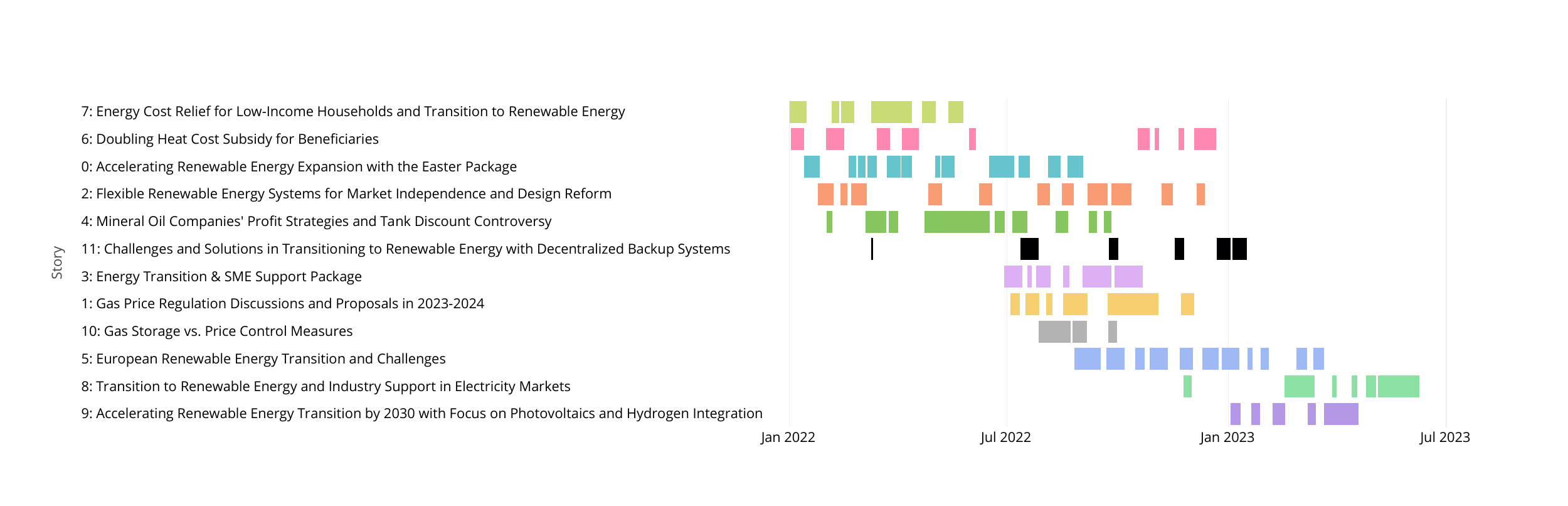}
    \caption{Extracted Stories with Events of Topic 11 (\textit{energy, gas, renewable, electricity, price}). Each row represents a story. Bars in each story correspond to events, whose width is defined by the events' start and end dates.}
    \label{fig:stories}
\end{figure}

To create coherent storylines, we leverage the events similarity matrix $S_2$. 
We construct a directed graph $G = (V, E)$, where $V$ represents the set of events and $E$ the edges between them.
Each edge $e_{ij}$ between events $i$ and $j$ is assigned a weight based on their similarity score $S_2(i, j)$.
The edge direction is determined by the chronological order of the events given by the representative documents, ensuring that each event has a pathway to all subsequent events.
To maximize coherence, we retain only the highest-weighted outgoing edge for each event.
This helps us focus on the main storyline avoiding diverging substories.
Finally, to separate the storylines, we apply the Leiden algorithm \cite{traag2019louvain} for community detection, with each partition representing a different story.

Figure \ref{fig:stories} illustrates the outcome of the story extraction process for topic 11 (\textit{energy, gas, renewable, electricity, price}), which resulted in the creation of 12 stories from 155 events.

\subsection{Separation of Narratives}
\label{sec:separation-of-narratives}
To distinguish the different viewpoints expressed in the extracted stories, we aim to identify users who share similar stances and are likely to contribute to the same narratives.
Since politicians often hold aligned opinions on various topics, such as party policies, we adopt a global (i.e., not restricted to one particular topic) approach to user grouping.
This method further ensures that even users less active within a given topic are appropriately categorized based on their overall trace of tweets.

Using a user's posts, we aim to create an embedding that captures their overall position.
Text embeddings have been shown to successfully quantify the degree of political bias in texts~\cite{DBLP:conf/www/GordonBM20} and identify ideological placement~\cite{Rheault_Cochrane_2020, elejalde2018nature}.
Moreover, they can highlight differences in how political groups discuss specific policy issues~\cite{Fredén04122024}.
To determine each user's overall stance, we first average the document embeddings of their posts within each single topic.
This gives us a topic-specific representation of the user's perspective.
Next, we calculate the user's relative position within that topic by subtracting the topic centroid from the user's average embedding.
This step helps to situate the user's stance relative to the broader discourse on that issue.
Finally, we average these topic-specific user embeddings across all topics.
This global embedding represents the user's overall political stance across various issues.

To unravel the diverse perspectives shaping the narratives within a story, we use the commonalities of users involved in terms of their global embeddings.
For this, we cluster the user embeddings (via HDBSCAN) to form distinct communities of users who share comparable political views.
With multiple users contributing to a story, we can now split the story posts according to their respective communities.
As a result, each event within a story is enriched by the perspectives of the various user communities, reflecting the collective voices that contribute to the discourse.
\textit{The structure represented by a given community's contribution to one particular story constitutes a competing narrative within that story.}

\section{Results}
We present the results based on applying the proposed framework to the dataset of German politicians' tweets.
We first evaluate the division of user positions to assess the overall coherence and consistency of the narratives within each user community.
Then, we illustrate, through two examples, how the same story or event can be differently perceived and discussed by opposing groups.
Our analysis shows how the proposed framework can help reveal distinct narratives emerging from the political discourse. 
We validate our approach by showing how supporters of different (politically opposed) parties emphasizing divergent interpretations of the same events are segregated by our method into separate narratives.

\subsection{Evaluation of Uncovered User Communities}
\begin{figure}[ht]
    \centering
    \begin{subfigure}{0.49\textwidth}
        \centering
        \includegraphics[width=\linewidth,trim={9.5mm 7.5mm 4mm 25mm },clip]{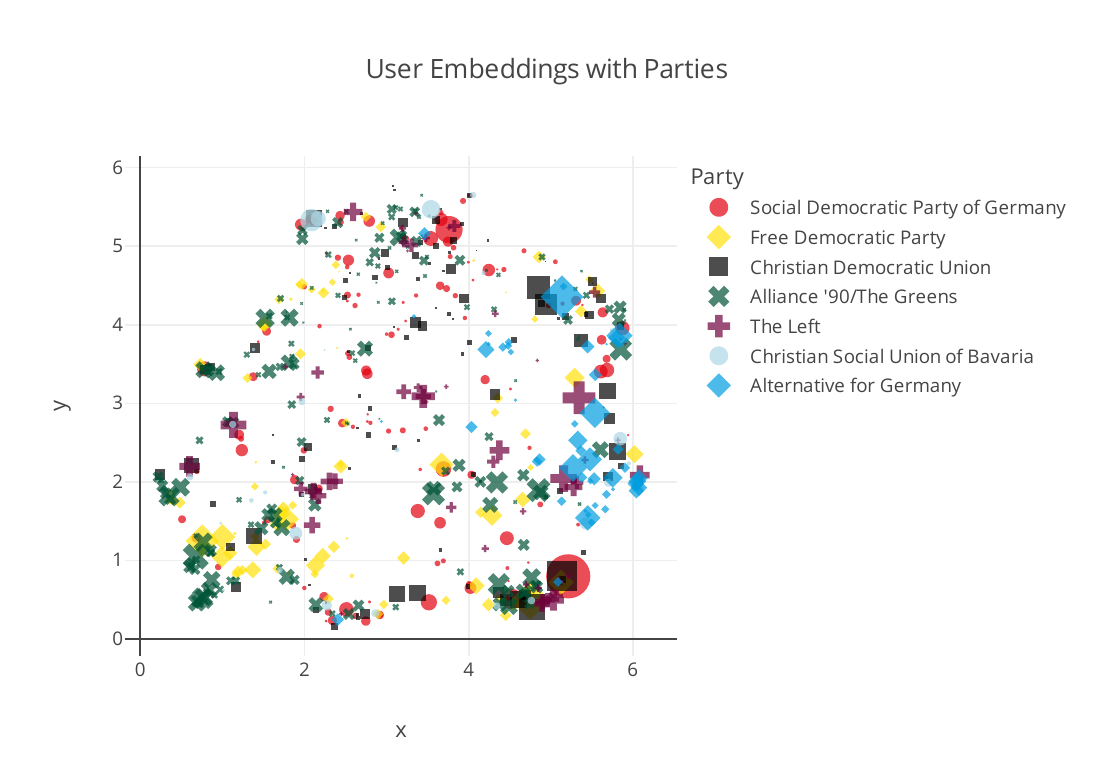}
        \caption{User embeddings colored by political party. Markers' size represents the users' activity frequency.\\}
        \label{fig:user-embeddings-party}
    \end{subfigure}
    \hfill%
    \begin{subfigure}{0.46\textwidth}
        \includegraphics[width=\linewidth,trim={9mm 6.5mm 20mm 29mm },clip]{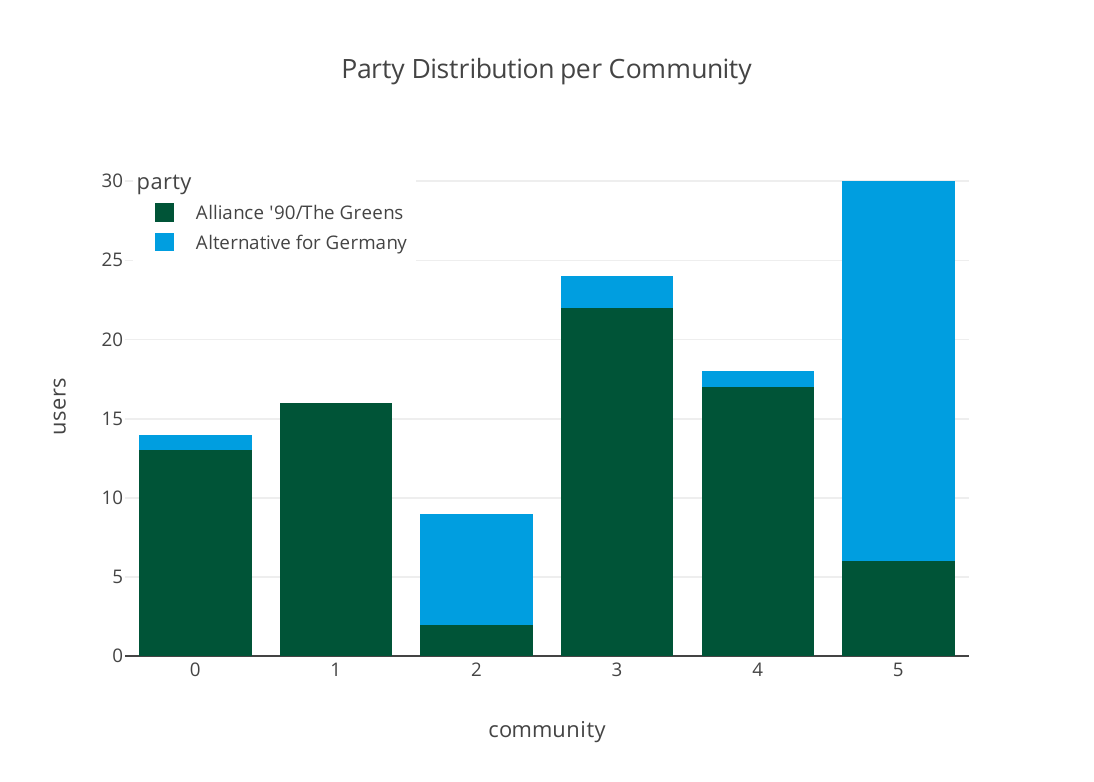}
        \caption{Distribution of members from two politically opposed parties within the user communities in Topic 23 (\textit{asylum, refugees, migration}).}
        \label{fig:community-parties}
    \end{subfigure}
    \caption{Forming Political Communities with Similar Stances through Global User Embeddings.}
\end{figure}

Since the identification of distinct narratives within each story relies on user communities, we first verify the validity of the user classification.
In our dataset, each user represents a German politician. 
Using Wikidata, we extracted information about their current party memberships.
This allows us to assess the alignment of the identified communities with their respective political parties.

Figure~\ref{fig:user-embeddings-party} shows a 2D projection of the user embeddings generated in Section~\ref{sec:separation-of-narratives}.
We focus our analysis on the main parties currently represented in the German parliament with 758 users in our dataset.
The figure indicates a good characterization of politicians based on their party affiliations.
For example, \sethlcolor{partygreens!40}\hl{The Greens} party members form a cluster mostly to the left side of the embedding space, reflecting their shared ideological positions.
In stark contrast, the right-wing Alternative for Germany (\sethlcolor{partyafd!50}\hl{AfD}) party occupies a cluster clearly separated and opposed to the Greens, highlighting the deep ideological divides between these two political factions.
Other major parties, such as the \sethlcolor{partycdu!30}\hl{CDU}/\sethlcolor{partycsu!70}\hl{CSU} and \sethlcolor{partyspd!50}\hl{SPD}, are positioned more centrally, with their members distributed across the embedding space in a manner that also corresponds to their relative ideological stances.
Further, we can observe an alignment of vocal members of the liberal party \sethlcolor{partyfdp!50}\hl{FDP} with some of the Greens positions (bottom left).

The generated user communities (see Section \ref{sec:separation-of-narratives}) should express different viewpoints within each topic.
We do not expect a clean separation of the individual parties given the heterogeneous views present within the larger, more central parties.
Instead, for validation purposes, we focus our analysis on parties with relatively polarized and objectively opposing positions, namely \sethlcolor{partygreens!40}\hl{The Greens} and the \sethlcolor{partyafd!50}\hl{AfD}.
These two parties appear well-separated within the identified communities, enabling the downstream use for narrative separation.
Figure~\ref{fig:community-parties} exemplifies the six identified communities and their party affiliations composition for topic 23 (\emph{asylum, refugees, migration}) when considering the two selected parties.
Note, however, as implied by the embeddings overview, that these groups additionally include other members from multiple other parties, covering the sizeable political landscape.

\subsection{Narratives Around Germany's 2022 Energy Crisis}
\begin{figure}[t]
    \centering
    \begin{subfigure}{0.6\textwidth}
        \centering
        \includegraphics[width=\linewidth]{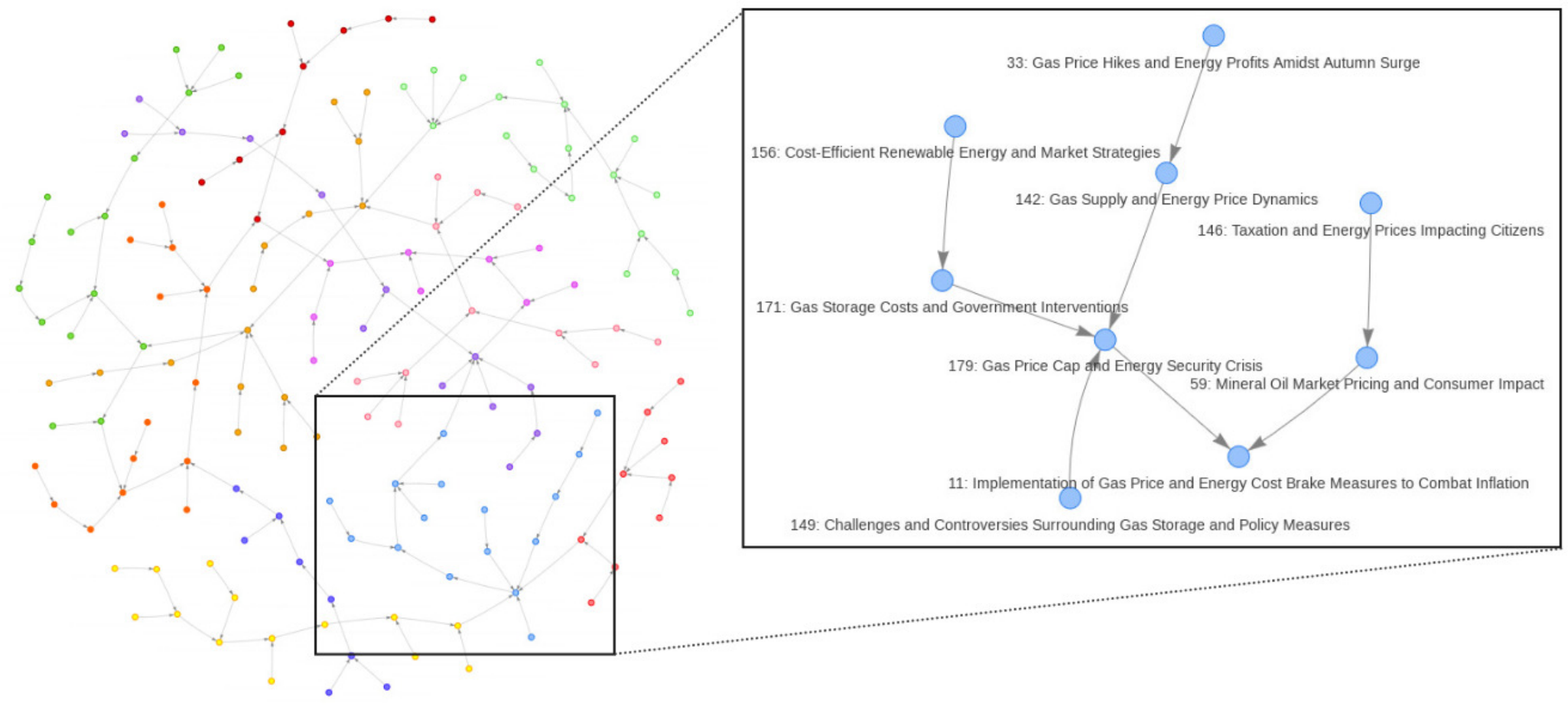}
        \caption{Story 10 within the directed graph of events}
        \label{fig:story-graph}
    \end{subfigure}
    \hfill
    \begin{subfigure}{0.39\textwidth}
        \includegraphics[width=\linewidth,trim={8mm 7mm 3mm 25mm },clip]{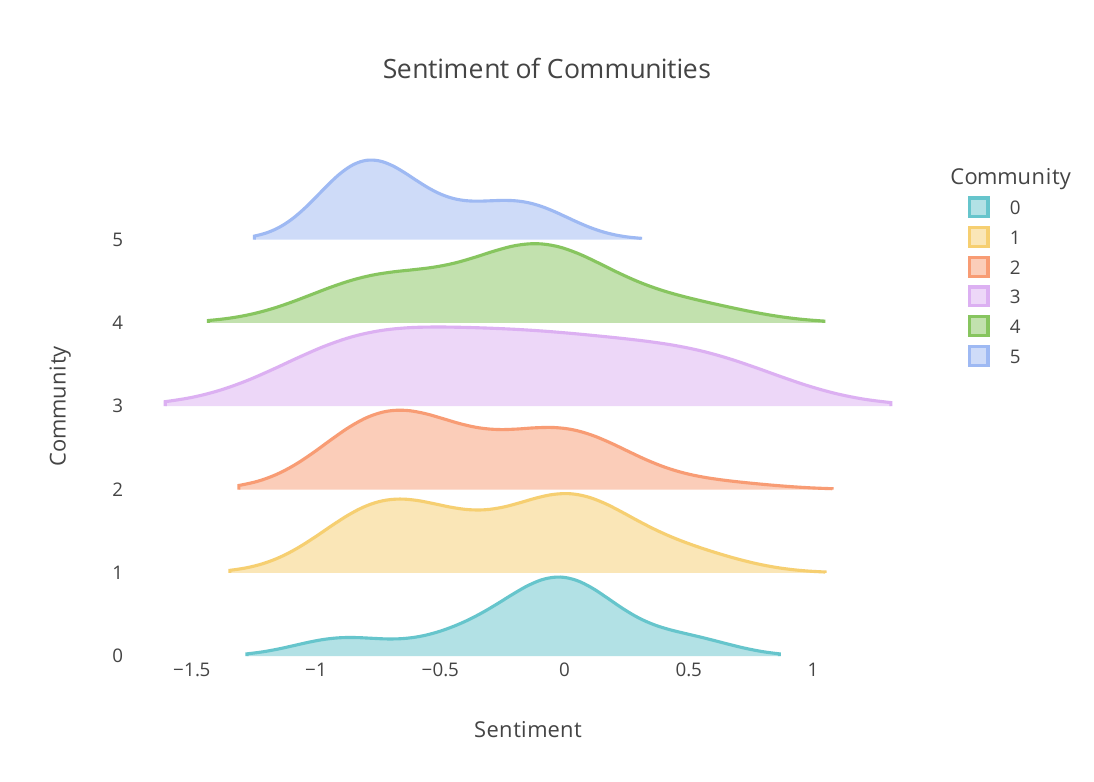}
        \caption{Sentiment Distribution of Communities in Story 10}
        \label{fig:story-sentiment}
    \end{subfigure}
    \caption{Story 10 (\textit{Gas Storage vs. Price Control Measures}) in Topic 11 (\textit{energy, gas, renewable, electricity})}
\end{figure}

As an insightful example for competing narratives, we focus on the discussions surrounding the energy crisis in Germany due to gas supply issues with Russia in 2022, which is captured by Topic 11 (\textit{energy, \dots}).
From the story graph in Figure \ref{fig:story-graph}, we select the debate over \enquote{\emph{Gas Storage vs. Price Control Measures}}.
This story links the discussion of a planned gas price surcharge due to higher purchase costs with the later adopted \enquote{gas price cap} by the ruling \enquote{traffic light} (Ampel) coalition.
Further references are made to increased storage costs, related taxation, and the excess profits of energy companies.

To delineate the differences in the viewpoints of the extracted communities and to identify contrasts, we first use sentiment analysis. 
Figure \ref{fig:story-sentiment} shows the sentiment distribution within them, with documents classified using XLM-T~\cite{barbieri-etal-2022-xlm}.
Comparing the dominant communities (i.e., most posts in the story), \sethlcolor{com1!50}\hl{1} and \sethlcolor{com5!50}\hl{5}, the latter uses noticeably more negative language, suggesting different framing.

Further, large language models allow us to summarize the community voices in order to demonstrate tangible differences in storytelling.
The dominant positions taken by both communities are summarized in Figure \ref{fig:community-voices-topic-11}.
Here, we prompt the phi-4 model \cite{abdin2024phi4technicalreport} for all posts per community, gaining insights without sifting through the content.

In the ongoing debate over rising energy costs, the two communities find common ground in their advocacy for financial relief for households and their opposition to the proposed surcharge, but their narratives diverge as they each shed light on different matters.
\sethlcolor{com1!50}\hl{Community 1} supports government intervention to relieve the economy and also mentions long-term strategies such as the transition to renewable energy or, more radically, the nationalization of energy companies and the lifting of the debt brake.
In contrast, \sethlcolor{com5!50}\hl{Community 5} is mainly skeptical about the current government's approach and questions its intentions, but fails to mention any far-reaching solutions. 
The negative sentiment observed is also reflected in the individual posts, which are characterized by strong language. 
For example, the crisis is attributed a purely political origin and the main players in the government are defamed, while the planned gas surcharge is described as \enquote{truly miserable}, unconstitutional and insane.

\begin{figure}
    \centering
    \small
\begin{tcolorbox}[skin=bicolor,size=title,colback=com1!30,colbacklower=com5!30,colframe=gray,width=0.95\linewidth]
    \emph{\enquote{Community 1 predominantly expresses urgent concern over rising energy costs, particularly focusing on gas prices, which have spurred calls for immediate relief measures such as the removal of the Gas Umlage (surcharge) and implementation of a Gas Price Cap or Break. There is consensus on the necessity for robust governmental intervention to mitigate economic strain caused by these escalating expenses. Many advocate for suspending fiscal constraints like the debt brake to facilitate necessary spending during this crisis, emphasizing that such financial flexibility is essential to provide timely support to households and businesses. Additionally, there's a strong demand for long-term solutions including transitioning away from fossil fuels towards sustainable energy alternatives, with some suggesting more radical actions like the nationalization of energy companies to secure public interest over corporate profits. Overall, while immediate relief measures are widely supported, there is also recognition of the need for strategic planning and investment in renewable technologies to address future challenges.}}
\tcblower
\emph{\enquote{Community 5 predominantly expresses strong opposition to the proposed Gas Surcharge, viewing it as economically burdensome, poorly conceived, and socially unfair. A dominant sentiment is that this measure was hastily implemented and crafted under influence from profit-driven entities rather than considering public welfare, highlighting a significant disconnect between government actions and citizens' interests. There's widespread support for the decision to halt the Gas Surcharge and replace it with measures like a Gas Price Brake, which are perceived as more equitable solutions aimed at alleviating the financial strain on households while ensuring economic stability without violating fiscal constraints such as the debt brake. The discourse also underscores frustration with governmental inefficiency and lack of transparency in decision-making processes, advocating for clearer communication and effective crisis management strategies that prioritize public interest over political or corporate gains. Overall, there is a call for more prudent and socially responsible policy-making that adequately addresses energy costs while safeguarding both individuals' financial well-being and broader economic health.}}
\end{tcolorbox}
    \caption{Generated summaries of community voices 1 and 5 within Story 10 (\textit{Gas Storage vs. Price Control Measures}) in Topic 11 (\textit{energy, gas, renewable, electricity})}
    \label{fig:community-voices-topic-11}
\end{figure}

\normalsize

\subsection{Law on Better Residence Opportunities for Migrants}
\begin{table}[ht]
    \caption{Diverse community positions within an Event in Story \enquote{Modern immigration law and civil equality in Germany} of Topic 23 (asylum, refugees, migration). [translated from German]}
    \label{tab:event-communities}
    \tiny
    \setlength{\tabcolsep}{2pt}
    \centering
    \begin{tabular}{cp{22mm}p{24mm}p{23mm}p{23mm}p{22mm}p{23mm}}
        \toprule
         \textbf{Community} & \sethlcolor{com0!50}\hl{\textbf{0}} & \sethlcolor{com1!50}\hl{\textbf{1}} & \sethlcolor{com2!50}\hl{\textbf{2}} & \sethlcolor{com3!50}\hl{\textbf{3}} & \sethlcolor{com4!50}\hl{\textbf{4}} & \sethlcolor{com5!50}\hl{\textbf{5}} \\
        \midrule
         \textbf{Central Post} & 
         \sethlcolor{com0!50}\hl{As \#Ampel, we stand for a paradigm shift in \#migration policy. We are opening up more legal ways to come to Germany and work here. Because we have a shortage of workers everywhere.} &
         \sethlcolor{com1!50}\hl{It is true. We need more regulated immigration and less illegal immigration. That includes: anyone who does not have a right of residence in Germany for reasons of political persecution or humanitarian emergency cannot stay in the country.} &
         \sethlcolor{com2!50}\hl{German SMEs are so desperately in need of new employees that your concerns fade into the background. Of course, we must consistently prevent migration into the social systems. But our country must be a country of immigration.} &
         \sethlcolor{com3!50}\hl{The Bundestag has just debated our new \#ResidenceOpportunitiesAct for the first time. Tolerated people who have been afraid of deportation for years now have prospects in Germany. Good for people, good for companies!} &
        \sethlcolor{com4!50}\hl{The Residence Opportunities Act gives people who have lived in Germany for many years more planning security for training or a job.} &
        \sethlcolor{com5!50}\hl{We need a deportation offensive. \#Germany must no longer be a haven for mentally conspicuous "lone offenders". The \#security of citizens must have priority. We owe this to the many victims of migration policy since 2015.} \\
         \bottomrule
    \end{tabular}
\end{table}

In addition to unpacking the community positions in the overall context of a story, we want to address the different facets of an event.
For this example, we switch to the polarizing debate on migration in Topic 23, where the planned facilitation of the citizenship of migrant workers is discussed. 

Table \ref{tab:event-communities} shows the extracted representative tweets per community within an event. 
The perspectives on work migration and swifter integration are divided into generally positive statements from Communities \sethlcolor{com0!50}\hl{0}, \sethlcolor{com3!50}\hl{3}, and \sethlcolor{com4!50}\hl{4}, positive with limitations from \sethlcolor{com1!50}\hl{1} and \sethlcolor{com2!50}\hl{2}, and general rejection from Community \sethlcolor{com5!50}\hl{5}.
The main similarities are the recognition of the need for more regulated immigration and the desire to address the labor shortage in Germany.
However, their approaches differ, with some advocating for more open and inclusive policies, while others prioritize security and deportation.
Different framings are employed to support the narratives.
For example, Communities \sethlcolor{com0!50}\hl{0} and \sethlcolor{com2!50}\hl{2} frame the issue in terms of economic needs, while Communities \sethlcolor{com1!50}\hl{1} and \sethlcolor{com5!50}\hl{5} focus on security and preventing illegal immigration.
Conversely, Communities \sethlcolor{com3!50}\hl{3} and \sethlcolor{com4!50}\hl{4} highlight the positive impact of the new Residence Opportunities Act.

Dissecting community involvement within each event in a story can also give analysts insights into the narrative framing at a more structural level. For example, we can track which events different narratives try to emphasize by looking at their community distribution and which events communities select to skip altogether. Moreover, analysts can study the timing and order in which competing narratives cover the events in the story to try to uncover internal dynamics and driving factors.

\section{Conclusions and Future Work}

This work introduces an unsupervised framework for identifying and analyzing competing political narratives on social media, focusing on German politicians' tweets. 
We uncovered competing narratives by automatically organizing topics and events into coherent stories and leveraging politician embeddings to identify distinct community perspectives.
We validate our method through illustrative examples related to the Energy Crisis and Migration Policy, demonstrating the framework's effectiveness in separating competing narratives within highly polarized political discourse.
This approach to large-scale social media analysis offers valuable insights, underscoring the potential of automatic narrative framing to monitor how political actors promote divergent interpretations of events on social media.
These findings provide valuable insights for policymakers, social media platforms, and researchers seeking to better understand political discourse, address polarization, and foster more balanced public discussions.

It is essential to highlight that the development of computational approaches for analyzing competing narratives relies on suitable datasets and evaluation metrics.
Existing datasets for narrative extraction often lack annotations related to competing narratives, making it challenging to train and evaluate models that can effectively distinguish and analyze them \cite{santana2023survey}.
Furthermore, developing robust metrics to assess the quality and coherence of extracted narratives and the accuracy of framing analysis remains an open challenge \cite{ranade2022computational, reiter2024computational}.
In this work, we rely primarily on qualitative evaluation.
However, we see potential to further strengthen our approach by correlating the linking of events to stories with the user streams.
This would allow us to measure and potentially enhance the coherence of the narratives extracted.
Initial experiments in this direction were promising, helping identify more coherent storylines that better reflect the framing and propagation of narratives across different political communities.

\section*{Acknowledgments}
We want to thank our former student Ahmad Hamadeh for the thoughtful discussions - his exploratory work laid an excellent foundation for further development.

\section*{Declaration on Generative AI} 
 During the preparation of this work, the author(s) used Claude 3 Haiku and Grammarly in order to: Grammar and spelling check, paraphrase, and reword. After using these tool(s)/service(s), the author(s) reviewed and edited the content as needed and take(s) full responsibility for the publication’s content. 

\bibliography{refs}

\end{document}